\documentclass{article}
\PassOptionsToPackage{numbers, compress}{natbib}

\usepackage[preprint]{neurips_2023}

\usepackage{subfigure}
\usepackage[utf8]{inputenc} 
\usepackage[T1]{fontenc}    
\usepackage{hyperref}
\usepackage{url}            
\usepackage{booktabs}       
\usepackage{amsfonts}       
\usepackage{nicefrac}       
\usepackage{microtype}      
\usepackage{bm}
\usepackage{amsmath, amssymb}
\usepackage{amsthm}
\usepackage[ruled,linesnumbered]{algorithm2e}
\usepackage{enumitem}
\usepackage{thm-restate}
\usepackage[table]{xcolor}
\usepackage{multirow}
\usepackage{graphicx}
\usepackage{tcolorbox}
\usepackage{titletoc}
\usepackage{soul}
\usepackage{algorithmic}
\usepackage{setspace} 
\usepackage{xcolor}
\usepackage{caption}
\usepackage{adjustbox}
\usepackage{diagbox}
\tcbuselibrary{skins}
\hypersetup{
	colorlinks=true,
	linkcolor=blue,
	citecolor=orange,
}

\definecolor{Top2}{RGB}{102, 171, 221}
\definecolor{Top1}{RGB}{245, 137, 112}

\title{P2RBox: Point Prompt Oriented Object Detection with SAM}
\author{Guangming Cao$^{1*}$, Xuehui Yu$^{1}$\thanks{Equal contribution.} , Wenwen Yu$^{1}$, Xumeng Han$^{1}$, \textbf{Xue Yang}$^{2}$ \\ \textbf{Guorong Li}$^{1}$, \textbf{Jianbin Jiao}$^{1}$, \textbf{Zhenjun Han}$^{1}$\thanks{Corresponding author.}\\
$^{1}$University of Chinese Academy of Sciences \quad 
$^{2}$Shanghai AI Laboratory \\
\small\texttt{caoguangming21@mails.ucas.ac.cn} \\
}

\newcommand{\ie}{\textit{i}.\textit{e}.}

\newcommand{\modelname}{P2RBox}

\begin{document}

\maketitle

\begin{abstract}
Single-point annotation in oriented object detection of remote sensing scenarios is gaining increasing attention due to its cost-effectiveness.
However, due to the granularity ambiguity of points, there is a significant performance gap between previous methods and those with fully supervision.
%
In this study, we introduce P2RBox, which employs point prompt to generate rotated box (RBox) annotation for oriented object detection.
P2RBox employs the SAM model to generate high-quality mask proposals. These proposals are then refined using the semantic and spatial information from annotation points. The best masks are converted into oriented boxes based on the feature directions suggested by the model. P2RBox incorporates two advanced guidance cues: Boundary Sensitive Mask guidance, which leverages semantic information, and Centrality guidance, which utilizes spatial information to reduce granularity ambiguity. This combination enhances detection capabilities significantly.
To demonstrate the effectiveness of this method, enhancements based on the baseline were observed by integrating three different detectors.
Furthermore, compared to the state-of-the-art point-annotated generative method PointOBB, P2RBox outperforms by about 29\% mAP (62.43\% vs 33.31\%) on DOTA-v1.0 dataset,
which provides possibilities for the practical application of point annotations.

\end{abstract}


\section{Introduction}
Aerial object detection focuses on identifying objects of interest, such as vehicles and airplanes, on the ground within aerial images and determining their categories. With the increasing availability of aerial imagery, this field has become a specific yet highly active area within computer vision ~\cite{ren2015faster,lin2017focal,tian2019fcos,ding2019learning,xie2021oriented}. 
Nevertheless, obtaining high-quality bounding box annotations demands significant human resources. 
Weakly supervised object detection ~\cite{bilen2016weakly,tang2017multiple,tang2018pcl,chen2020slv,wan2018min,zhou2016learning,diba2017weakly,zhang2018adversarial} has emerged as a solution, replacing bounding box annotations with more affordable image-level annotations.

However, due to the absence of precise location information and challenges in distinguishing densely packed objects, image-level supervised methods~\cite{tan2023wsodet} exhibit limited performance in complex scenarios. 
Recently, the field of weakly supervised remote sensing has aspired to achieve a balance between performance and annotation cost by introducing slightly more annotation information, such as horizontal bounding box-based~\cite{yang2022h2rbox,yu2024h2rbox} and single point-based~\cite{yi2023point2rbox,luo2023pointobb} oriented object detection.

Horizontal bounding box-based methods have made significant progress, while point-supervised methods still exhibit a substantial performance gap compared to rotated bounding box supervision.

Given that SAM~\cite{kirillov2023segment} currently serves as a strong visual model, it possesses robust zero-shot capabilities and supports the use of points as prompt inputs.
A natural idea is to utilize point inputs to feed into the SAM to generate mask proposals, which are then converted into rotated bounding box annotations.
This approach introduces a new challenge: selecting the most appropriate mask proposal among the many generated by SAM.
Although the SAM model assigns a initial score to each mask upon generation, our experiments indicate that this score frequently fails to accurately represent mask quality due to the lack of intra-class homogeneity, as demonstrated in Fig.~\ref{fig:motivation}.
Consequently, this paper introduces the P2RBox network, which utilizes the semantic and spatial information of the annotation points to develop a novel criterion for assessing mask quality.

In the P2RBox framework, a boundary-aware mask-guided MIL module and a centrality-guided module have been designed, which are utilized for the network to learn semantic and positional information respectively. These modules also facilitate the evaluation of mask quality based on the learned information. 
Additionally, to address the inevitable orientation issues in remote sensing datasets, particularly the ambiguity between the axis of symmetry of common targets such as airplanes and the orientation of the minimum bounding rectangle, P2RBox incorporates an angle prediction branch embedded in a centrality-guided module.
This predicted angle is tasked with transforming the generated mask proposals into oriented bounding rectangles.
The masks selected based on this new criterion show significant improvement and high-quality pseudo annotations are generated therefore. Our main contributions are as follows:
\begin{figure}[t]
\begin{center}
\includegraphics[width=0.92\textwidth]{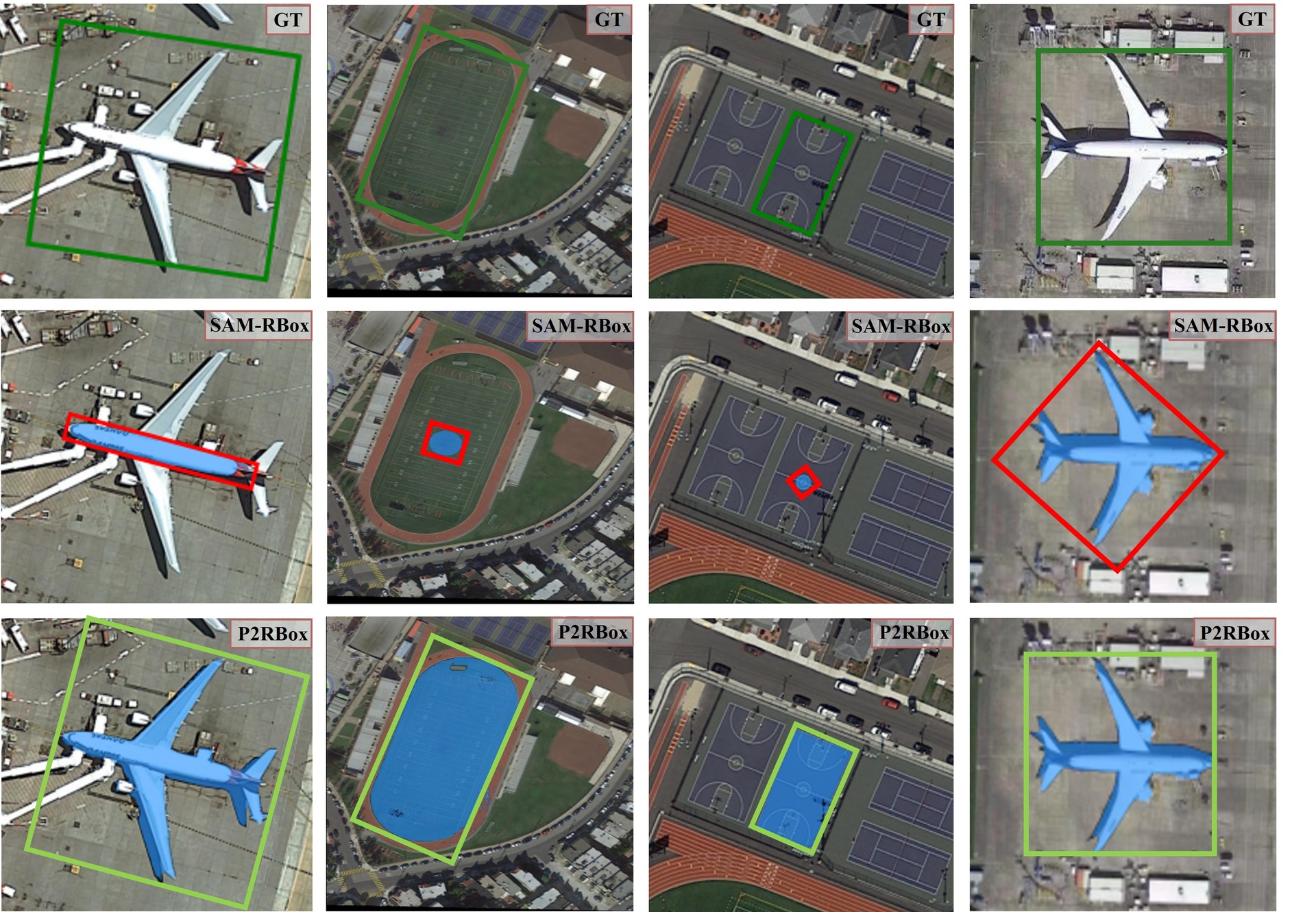}
\vspace{-3pt}
\caption{Visual comparison of the highest confidence mask and its corresponding rotated box annotation generated by mask proposal generator (SAM) and P2RBox. The second row displays the results of the baseline method, while the Rotated boxes in the last row are generated by the SAE module. (Best viewed in color).
}
\vspace{-18pt}
\label{fig:motivation}
\end{center}
\end{figure}

\textbf{1)} Proposing of the P2RBox Network: We introduce the P2RBox network, which is the first method based on point annotation and SAM for achieving point-supervised rotated object detection. By combining with Oriented R-CNN, P2RBox achieves 62.43\% on DOTA-v1.0 test dataset, which narrows the gap between point-supervised and fully-supervised methods.

\textbf{2)} 
High-Quality Mask Selection: By employing the boundary-aware mask-guided MIL module, we introduce a semantic score for the masks. This score, combined with the spatial offset score from the centrality-guided module, forms a comprehensive filtering approach. This approach enhances the quality of the selected mask proposals from the mask generator, ultimately improving the quality of the rotated bounding box annotations generated through the predicted angle.


\section{Related work}
\textbf{Weakly-supervised oriented object detection.} 
In the field of remote sensing, detectors utilizing rotated bounding box annotations have demonstrated robust performance and widespread applications~\cite{lin2017focal,tian2019fcos,ding2019learning,xie2021oriented,yang2021r3det,han2021align,han2021redet}. Concurrently, the use of weakly supervised annotations as a cost-effective method for reducing labeling expenses is gaining increasing attention.
OAOD~\cite{iqbal2021leveraging} is proposed for weakly-supervised oriented object detection. But in fact, it uses HBox along with an object
angle as annotation, which is just “slightly weaker” than
RBox supervision.
H2RBox ~\cite{yang2022h2rbox}directly detects RBox by using HBox annotations, leveraging geometric constraints to enhance detection efficiency. H2RBox-v2~\cite{yu2023h2rbox} further introduces the reverse angle sub-supervision module, which significantly improves the performance and is equivalent to the fully supervised detector. PointOBB~\cite{luo2023pointobb} is the first method using single-point supervision for oriented object detection, employing self-supervised learning for object scale and angle with innovative loss strategies. Point2RBox~\cite{yi2023point2rbox} introduces a lightweight, end-to-end model that uses synthetic patterns and self-supervision to predict rotated bounding boxes from point annotations efficiently

\textbf{Semi-supervised oriented object detection.}
Semi-supervised learning sits between supervised and unsupervised learning, utilizing a small amount of labeled data in conjunction with a large volume of unlabeled data for training~\cite{zhu2023knowledge, hua2023sood, zhou2023semi,wu2024pseudo}.
SOOD~\cite{hua2023sood} designed a semi-supervised oriented object detection method for remote sensing images. This method is based on a pseudo-label framework and introduces two novel loss functions: Rotation Aware Weighted Adaptation (RAW) loss and Global Consistency (GC) loss, to optimize the detection of oriented targets. 
Zhou et al.~\cite{zhou2023semi} proposed an end-to-end semi-supervised SAR ship detection framework with Interference Consistency Learning (ICL) and Pseudolabel Calibration Network (PLC) to improve accuracy and robustness against complex backgrounds and interferences in SAR imagery.



\textbf{Enhancements and Applications of SAM.} In the rapidly evolving field of computer vision, the Segment Anything Model (SAM)~\cite{kirillov2023segment} has emerged as a versatile tool, capable of generating object masks from simple user inputs~\cite{chan2024sam3d, zohranyan2024dr,li2024concatenate, liu2024wsi,chen2024sam}, such as clicks. SAPNet~\cite{wei2023semantic} that integrates SAM with point prompts to tackle the challenge of semantic ambiguity, utilizing Multiple Instance Learning (MIL) and Point Distance Guidance (PDG) to refine SAM-generated masks, ensuring high-quality, category-specific segmentation. UV-SAM~\cite{zhang2024uv} aims to identify urban village boundaries in cities by adjusting SAM. The research uses the hybrid cues generated by the semantic segmentation model to be input into SAM to achieve fine-grained boundary recognition.
ASAM~\cite{li2024asam}enhances SAM through adversarial tuning, using natural adversarial examples to improve segmentation tasks significantly. This method requires no extra data or changes to the architecture, setting new performance benchmarks.
Dual-SAM~\cite{zhang2024fantastic} enhances SAM for marine animal segmentation by introducing a dual structure to improve feature learning. It incorporates a Multi-level Coupled Prompt (MCP) strategy for better underwater prior information and a Criss-Cross Connectivity Prediction (C3P) method to enhance inter-pixel connectivity. 

Although this paper uses point annotations to generate oriented bounding boxes, unlike traditional networks, it employs SAM to generate high-quality candidate masks. Therefore, this paper not only utilizes point annotations but also extracts valuable information from them to optimize the selection of candidate masks.
\begin{figure}[htbp]
\begin{center}
\includegraphics[width=0.95\textwidth]{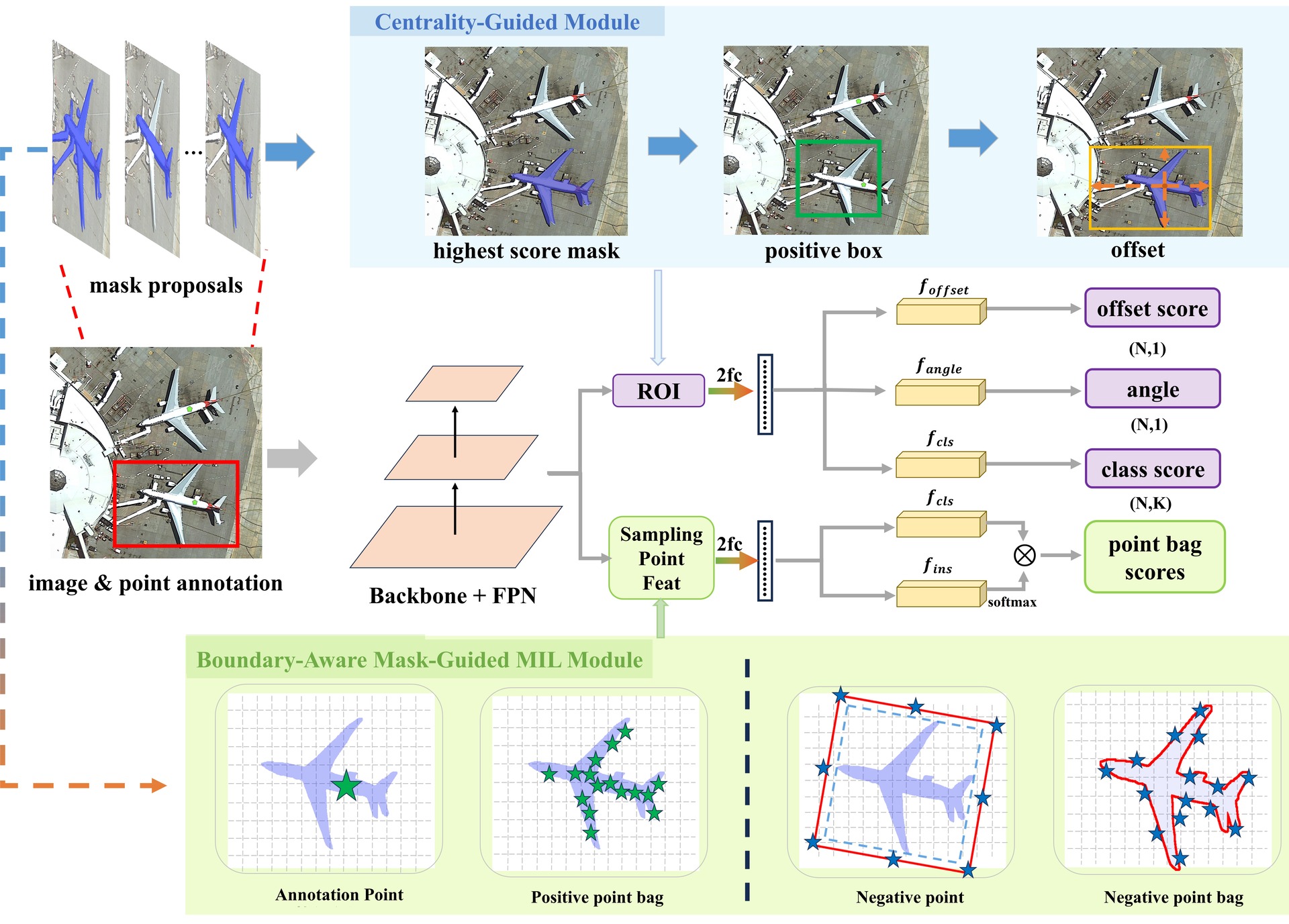}
\vspace{-2pt}
\caption{The overview of training process of {\modelname}, consisting of boundary-aware mask-guided MIL module, centrality-guided module.
Initially, mask proposals are generated by a generator (SAM). Four point sets are constructed according the highest initial score mask to train boundary-aware mask-guided MIL moudle, which pursuing dataset-wide category consistency.
The centrality-guided module generates offset penalty to enhance alignment between points and masks.
The trained network will used to assesses mask quality, selecting the best proposals for detector training supervision. (Best viewed in color)
}
\vspace{-10pt}
\label{fig:framework}
\end{center}
\end{figure}
\section{Method}

\subsection{Framework}
The generative network structure allows for the conversion of annotations and the replacement of strong supervised detectors to obtain different models with varying focuses (e.g., performance or detection speed). This maintains a certain degree of flexibility while achieving good performance.
As shown in the Fig ~\ref{fig:framework}, this work designs the P2RBox network, which inputs annotation points into the SAM model. By establishing an evaluation mechanism, the filtered masks are retained and converted into rotated box annotations. This process is used to train a strongly supervised rotated object detector, accomplishing the transition from point annotations to rotated object detection.

The evaluation mechanism established in this paper is the key focus of our work. We designed two modules to separately mine the feature information of annotation points and RoIs, and developed evaluation criteria.
The first module is the boundary-aware mask-guided MIL module. In the neighborhood of the annotation points, it constructs positive point bags, negative point bags, and negative points to guide the model in learning the class information of the points on the feature map.
The second module is the centrality guided module. Based on the regions of interest corresponding to the candidate masks, it extracts features and uses them to predict the spatial offsets of the masks.

In the inference part of the P2RBox, the output results from these two modules will be used to design a comprehensive mask evaluation metric.
In addition, to convert masks into rotated boxes, based on RoI features, this paper designs an angle prediction branch to guide the generation of oriented bounding rectangles from the masks.
During the training process of the network, only the class information is known and can be directly used to calculate the loss.
The other two important branches, such as spatial offset and angle prediction, both use calculated values based on the algorithm designed in this paper as supervision. During inference, the trained network predictions are used.

\subsection{Boundary-Aware Mask-Guided MIL Module}
The information contained in point annotations is merely the class information of a single point.
After generating candidate masks using the SAM model with annotation point prompt, different methods for generating point bags based on these candidate masks were designed in this paper.
Given a candidate mask, a positive point set corresponding to the candidate mask is constructed by randomly sampling pixel points within the mask.
This sampling method not only utilizes the information from the mask but also, with a sufficient number of samples, can represent this type of mask.
\begin{equation}
    \mathcal{B}_{pos}=\big\{p_i|p_i \in mask,\ p_i\ is\ randomly\ selected\big\},
\end{equation}

Background points located far from the object are easier for the network to learn. Conversely, the negative points along the object's boundary are more valuable samples, as they often reside on the critical boundary between the foreground and background.
Hence, a negative point bag is designed to train the model to discern boundary features. By selecting several points in the margin pixels of the mask, the negative point bag can be defined as:
\begin{equation}
    \mathcal{B}_{neg}=\big\{p_i|p_i \in mask_{margin},\ p_i\ is\ randomly\ selected\big\},
\end{equation}

With annotated point $a$ as a naturally positive reference, we introduce negative points to instruct the model in discerning background features.
To obtain the negative points for a given mask, we follow a three-step process. Firstly, determine the circumscribed bounding box of the mask, denoted as $(x,\,y,\,w,\,h,\,\alpha)$.
Secondly, increase both the height $h$ and width $w$ by a factor as $\hat{h}=(1+\delta)\cdot h$ and $\hat{w}=(1+\delta)\cdot w$, where $\delta$ is the set to $0.05$ in this paper.
Lastly, the set of negative points, denoted as $\mathcal{N}$, comprises the four vertices along with the midpoints of the four edges, \ie,
\begin{equation}
\begin{split}
    n_{ij} = (x+\frac{\hat{w}}{2}\cdot\cos\alpha\cdot i-\frac{\hat{h}}{2}\cdot\sin\alpha\cdot j &,\, y+\frac{\hat{w}}{2}\cdot\sin\alpha\cdot i+\frac{\hat{h}}{2}\cdot\cos\alpha\cdot j), \\
    \mathcal{N}=\big\{n_{ij}\,|\,i,j\in\{-1,0,1&\},\,(i,j)\neq(0,0)\big\}.
\end{split}
\end{equation}
To facilitate \modelname\ in determining whether the points in $\mathcal{B}$ belong to the same category as annotated point, we treat the points in bag $\mathcal{B}$ as positive instances. We extract the feature vectors $\{\mathbf{F}_p | p \in \mathcal{B} \}$. For each $p \in \mathcal{B}$, two separate fully connected layers with independent weights generate the classification score and instance score, denoted as $[S_{\mathcal{B}}^{ins}]_p$ and $[S_{\mathcal{B}}^{cls}]_p$.

To obtain the final classification score for $\mathcal{B}$, we calculate it by summing the element-wise products of $[S_{\mathcal{B}}^{ins}]_p$ and $[S_{\mathcal{B}}^{cls}]_p$ as follows:
\begin{equation}
S_{\mathcal{B}} = \sum_{p \in \mathcal{B}} [S_{\mathcal{B}}^{ins}]_p \cdot [S_{\mathcal{B}}^{cls}]_p, \in \mathbb{R}^{K}.
\end{equation}

For annotation points or negative points, directly use the prediction results from the classification score branch.
In the inference stage, the positive point bag and the negative point bag will be used together to evaluate the mask.

\subsection{Centrality-Guided Module}
For generative point-based approaches like P2BNet~\cite{chen2022point} and PointOBB~\cite{luo2023pointobb}, they generate anchor boxes centered on annotated points, thereby utilizing the points' positional information.
Utilizing semantic information, P2RBox has demonstrated the ability to discern the quality of candidate masks. However, existing modules mainly focus on extracting semantic information and fail to fully leverage the spatial information provided by point annotations. This deficiency often leads to generated masks containing redundant areas when covering the target.

The centrality-guided module also operates based on feature maps and candidate masks, as shown in Fig.~\ref{fig:framework}.
The specific implementation steps are as follows: 
1) For the mask, alculate the coordinates of its bounding horizontal box $(x_1,y_1,x_2,y_2)$, where $(x_1,y_1)$ is the top-left corner and $(x_2,y_2)$ is the bottom-right corner. 2) Sample features from the feature map based on the RoI to obtain the corresponding feature vector. 
3) Through two fully connected layers with shared weights, three different branches are designed. The first is the offset prediction branch, which outputs an $N\times 1$ vector, where $N$ is the number of targets in the image, and 1 represents the offset prediction for each target. The second is the classification branch, used to calculate the class score of the region of interest. The last is the angle prediction branch, which outputs an $N\times 1$ vector.

In this thesis, a benchmark ground truth value for supervised learning was designed for the network's offset prediction , and a reference value $gt_{offset}$ to guide the model training.
Based on the coordinates of the annotation point $(x,y)$ the reference value for offset is designed as follows:
\begin{equation}
    gt_{offset}=1-\sqrt{\frac{min(x-x_1,x_2-x)}{max(x-x_1,x_2-x)} \cdot \frac{min(y-y_1,y_2-y)}{max(y-y_1,y_2-y)}}.
\end{equation}

The design of $gt_{offset}$ ensures that its value decreases as the annotation point approaches the center of the target; conversely, its value increases as the annotation point moves away from the target center, serving as a penalty mechanism for the predicted offset.

The classification branch outputs an $N\times K$ vector ($K$ represents the number of categories) , where the category information is determined by the category of the labeled points. As for angle prediction, due to the lack of angle information, an algorithm is designed below to supervise the angle.

\textbf{Angle Prediction.} Rotated bounding boxes offer a more precise means of annotation and can also convey its orientation. 
Generally, for symmetrical objects, the annotation direction of the rotated bounding box often aligns with the direction of the axis of symmetry. However, the minimum enclosing rectangle cannot guarantee this alignment, as shown in the last column of Fig. \ref{fig:motivation}.
Assume there is an object, with $P$ being the set of its pixel points. Interestingly, the axis of symmetry of the object can be obtained by solving for the eigenvector of $P^T \cdot P$ after translating its center of mass to the origin.A simple proof will be given here.

In accordance with this condition, if the target exhibits symmetry along an axis passing through the origin, then there exists a rotation matrix, also referred to as an orthogonal matrix $R$, such that:
\begin{equation}
    P\cdot E=Q\cdot R, E=\left(\begin{matrix}1&0\\0&1\end{matrix}\right).
\end{equation}
Here, we have a matrix $R$ with dimensions $2\times 2$, representing a rotated matrix that aligns the axis of symmetry with the x-axis. Consequently, we can express $Q$ as follows:
\begin{equation}
    Q=\left(\begin{matrix}x_1&x_1 &\ldots &x_n&x_n\\y_1&-y_1&\ldots&y_n&-y_n\end{matrix}\right)^T.
\end{equation}
To find the matrix $R$, we multiply both sides of the above equation by its transpose, yielding:
\begin{equation}
    E^{T}\cdot P^{T}\cdot P\cdot E=R^T\cdot Q^T\cdot Q\cdot R.
\end{equation}
This further simplifies to:
\begin{equation}
    P^T\cdot P=R^T\cdot \Bigg(\begin{matrix}2\times \sum_{i=1}^{n}x_{i}^{2}&0\\0&2\times \sum_{i=1}^{n}y_{i}^{2}\end{matrix}\Bigg)\cdot R.
    \label{equ:pca}
\end{equation}
According to the spectral theorem for symmetric matrices in advanced algebra (proposed by Sylvester in the 19th century): Every real symmetric matrix can be orthogonally diagonalized.
Eq. \ref{equ:pca} demonstrates that $Q^T\cdot Q$ and $P^T \cdot P$ are similar matrices, with $R$ serving as the similarity transformation matrix and also the eigenvector matrix of $P^T \cdot P$ because $Q^T\cdot Q$ being a diagonal matrix. This confirms our assertion: The eigenvectors of the matrix $P^T\cdot P$ correspond to the object's symmetry direction and its vertical direction.

\subsection{Total Loss and Mask Quality Assessment}
\textbf{Total Loss}. The loss function of P2RBox includes classification loss, offset prediction loss, and angle prediction loss, marked as $\mathcal{L}_{CGM}$. Object-level MIL loss ($\mathcal{L}_{MIL}$) is introduced to endow \modelname\ the ability of finding semantic points around each annotated point. By combining information from similar objects throughout the entire dataset, it imparts discriminative capabilities to the features of that category, distinguishing between foreground and background. The boundary sensitive mask guided loss function of \modelname\ is a weighted summation of the three losses:
\begin{equation}
\mathcal{L}=\mathcal{L}_{MIL}+\mathcal{L}_{CGM}.
\end{equation}
\begin{equation}
\mathcal{L}_{MIL}=\mathcal{L}_{ann}+\mathcal{L}_{neg}+\mathcal{L}_{bag}^{pos}+\mathcal{L}_{bag}^{neg}.
\end{equation}
\begin{equation}
    \mathcal{L}_{CGM}=\mathcal{L}_{box}^{cls}+\mathcal{L}_{offset}+\mathcal{L}_{angle}.
\end{equation}
The four terms in $\mathcal{L}_{MIL}$ correspond to the four different components designed in the boundary-aware mask-guided MIL module, and their output scores are calculated using focal loss. The three terms in $\mathcal{L}_{CGM}$ correspond to the outputs of the three branches in the centrality guided module, with the classification loss $\mathcal{L}_{box}^{cls}$ calculated using focal loss. The angle loss $\mathcal{L}_{angle}$ is calculated using smooth-l1 loss according to the supervision value in the module, while the offset loss $\mathcal{L}_{offset}$ is calculated using cross entropy loss between the prediction and the supervision value.

\textbf{Mask quality assessment}.
During inference, several outputs from the above modules will be used to assess the mask.
The specific calculation method is as follows.
\begin{equation}\label{eq:score}
    Score=S_{SAM}+S_{bmm}+S_{cgm}.
\end{equation}
\begin{equation}
    S_{cgm}=S_{boxCls}-S_{offset},
\end{equation}
\begin{equation}
    S_{bmm}=S_{posBag}-S_{negBag}.
\end{equation}
Using the softmax activation function to activate positive instance bags, then obtaining the values corresponding to each class, resulting in $S_{posBag}$. For negative instance bags, applying the sigmoid activation function to activate the scores of negative bags, then selecting the highest class score as the penalty term $S_{negBag}$.
Similarly, the classification scores of the boxes $S_{boxCls}$ predicted by the centrality guided module are activated using softmax, then the corresponding class scores are subtracted by the network's predicted offsets to obtain the $S_{cgm}$.

During the testing phase, we straightforwardly select the mask with the highest $Score$ as the object's mask, which is subsequently converted into a rotated bounding box with predicted angle.

\section{Experiments}
\subsection{Datasets and Implement Details}
\textbf{DOTA~\cite{xia2018dota}.} There are 2,806 aerial images, 1,411 for training, 937 for validation, and 458 for testing,
as annotated using 15 categories with 188,282 instances in total. We follow the preprocessing in
MMRotate~\cite{zhou2022mmrotate}, i.e. the high-resolution images are split into 1,024 $\times$ 1,024 patches with an overlap of 200
pixels for training, and the detection results of all patches are merged to evaluate the performance. We use training and validation sets for training and the test set for testing. The detection performance is obtained by submitting testing results to DOTA’s evaluation server. 

\textbf{DIOR-R~\cite{dior-r}.}
DIOR-RBox, derived from the original aerial image dataset DIOR~\cite{dior}, has undergone reannotation using RBoxes instead of the initial HBox annotations. With a total of 23,463 images and 190,288 instances spanning 20 classes, DIOR-RBox showcases extensive intra-class diversity and a wide range of object sizes.

In the experimental section and ablation studies of this paper, mAP$_50$ and mIoU are utilized as the primary evaluation metrics.

\textbf{Single Point Annotation.} Since there are no dedicated point annotations on the DOTA and DIOR-R datasets, this paper directly uses the center points of the bounding boxes annotated with rotation angles as the point annotations for dataset utilization. However, the method proposed in this paper does not require such strict adherence to the exact positions of the annotated points.
 
\textbf{Training Details.} P2RBox predicts the oriented bounding boxes from single point annotations and uses the predicted boxes to train three classic oriented detectors (RetinaNet, FCOS, Oriented R-CNN) with standard settings. All the fully-supervised models are trained based on a single GeForce RTX 2080Ti GPU. Our model is trained with SGD~\cite{bottou2012stochastic} on a single GeForce RTX 2080Ti GPU. The initial learning rate is $2.5\times 10^{-3}$ with momentum 0.9 and weight decay being 0.0001. And the learning rate will warm up for 500 iterations.
Training P2RBox on a single GPU typically takes around 9 hours.
To demonstrate the effectiveness of the proposed method in our model, we designed a parameter-free rotation box annotation generator based on SAM, which directly retains the highest initial score mask and computes the minimum bounding rectangle to obtain the rotated bounding box.

\begin{table*}[tb!]
	\begin{center}
	\caption{Detection performance of each category on the DOTA-v1.0 test set}\label{table:dota}
		\renewcommand\arraystretch{1.2}		\resizebox{1\textwidth}{!}{
			\begin{tabular}{l|ccccc ccccc ccccc |c}
				\toprule
				\textbf{Method}  & \textbf{PL} & \textbf{BD} & \textbf{BR} & \textbf{GTF} & \textbf{SV} & \textbf{LV} & \textbf{SH} &
				\textbf{TC} & \textbf{BC} & \textbf{ST} & \textbf{SBF} & \textbf{RA} & \textbf{HA} & \textbf{SP} & \textbf{HC} & \textbf{mAP$_{50}$}  \\
    			\hline \multicolumn{17}{l}{\textbf{{rbox-level:}}}  \\ \hline
				RetinaNet~\cite{lin2017focal}&89.1&74.5&44.7&72.2&71.8&63.6&74.9&90.8&78.7&80.6&50.5&59.2&62.9&64.4&39.7&67.83\\
                FCOS~\cite{tian2019fcos}&88.4&75.6&48.0&60.1&79.8&77.8&86.7&90.1&78.2&85.0&52.8&66.3&64.5&68.3&40.3&70.78\\
                Oriented R-CNN~\cite{xie2021oriented}&89.5&82.1&54.8&70.9&78.9&83.0&88.2&90.9&87.5&84.7&64.0&67.7&74.9&68.8&52.3&75.87\\
                \hline \multicolumn{17}{l}{\textbf{{hbox-level:}}}  \\\hline
                H2RBox~\cite{yang2022h2rbox}&88.5&73.5&40.8&56.9&77.5&65.4&77.9&90.9&83.2&85.3&55.3&62.9&52.4&63.6&43.3&67.82\\
                H2RBox-v2~\cite{yu2023h2rbox}&89.0&74.4&50.0&60.5&79.8&75.3&86.9&90.9&85.1&85.0&59.2&63.2&65.2&70.5&49.7&72.31\\
				\hline \multicolumn{17}{l}{\textbf{{point-level:}}}  \\ \hline
				PointOBB(ORCNN)~\cite{luo2023pointobb}&28.3&70.7&1.5&64.9&68.8&46.8&33.9&9.1&10.0&20.1&0.2&47.0&29.7&38.2&30.6&33.31 \\
				Point2RBox-SK~\cite{yi2023point2rbox}&53.3&63.9&3.7&50.9&40.0&39.2&45.7&76.7&10.5&56.1&5.4&49.5&24.2&51.2&33.8&40.27 \\
				\hline \multicolumn{17}{l}{\textbf{{SAM-based training:}}}  \\ \hline
				RetinaNet~\cite{lin2017focal}&79.7&64.6&11.1&45.6&67.9&47.7&74.6&81.1&6.6&75.7&20.0&30.6&36.9&50.5&26.1&47.91\\  
                FCOS~\cite{tian2019fcos}&78.2&61.7&11.7&45.1&68.7&64.8&78.6&80.9&5.0&77.0&16.1&31.8&45.7&53.4&44.2&50.84\\
                Oriented R-CNN~\cite{xie2021oriented}&79.0&62.6&8.6&55.8&68.4&67.3&77.2&79.5&4.4&77.1&26.9&28.8&49.2&55.2&51.3&52.75\\
                \hline \multicolumn{17}{l}{\textbf{{P2RBox-based training:}}}\\\hline
                \rowcolor{gray!20} RetinaNet&86.5&66.2&15.7&64.8&70.9&56.3&76.2&83.6&45.5&80.1&34.0&38.4&38.2&55.7&37.0&56.66\\
                \rowcolor{gray!20} FCOS&87.8&65.7&15.0&60.7&73.0&71.7&78.9&81.5&44.5&81.2&41.2&39.3&45.5&57.5&41.2&59.04\\
                \rowcolor{gray!20} Oriented R-CNN&87.4&66.6&13.7&64.0&72.0&74.9&85.7&89.2&47.9&83.3&45.7&40.6&50.1&65.6&49.1&62.43\\
                \bottomrule
		    \end{tabular}   }
	\end{center}
	\vspace{-10pt}
\end{table*}

\begin{table*}[tb!]
    \centering
    \caption{Detection performance of each category on the DIOR-R test set}\label{table:dior}
    \resizebox{1.0\textwidth}{!}{
    \fontsize{12pt}{12.0pt}\selectfont
        \renewcommand\arraystretch{1.6}
        \setlength{\tabcolsep}{0.8mm}
        \begin{tabular}{l|cccccccccccccccccccc|c}
        \toprule
        \textbf{Method} & \textbf{{APL}} & \textbf{APO} & \textbf{{BF}}  &  \textbf{{BC}} & \textbf{BR} & \textbf{CH} & \textbf{ESA} & \textbf{ETS} & \textbf{DAM} & \textbf{GF} & \textbf{{GTF}} &  \textbf{{HA}} & \textbf{OP} & \textbf{{SH}} & \textbf{STA} & \textbf{STO} & \textbf{{TC}} &  \textbf{TS} &  \textbf{{VE}} & \textbf{WM} & 
        \textbf{mAP$_{50}$} \\ \hline
        \multicolumn{22}{l}{\textbf{rbox-level:}} \\ \hline
        RetinaNett~\cite{lin2017focal}& 58.9 & 19.8 & 73.1 & 81.3 & 17.0 & 72.6 & 68 & 47.3 & 20.7 & 74.0 & 73.9 & 32.5 & 32.4 & 75.1 & 67.2 & 58.9 & 81.0 & 44.5 & 38.3 & 62.6  & 54.96 \\
        FCOS~\cite{tian2019fcos} & 61.4 & 38.7 & 74.3 & 81.1 & 30.9 & 72.0 & 74.1 & 62.0 & 25.3 & 69.7 & 79.0 & 32.8 & 48.5 & 80.0 & 63.9 & 68.2 & 81.4 & 46.4 & 42.7 & 64.4  & 59.83 \\ 
        Oriented R-CNN  & 63.1 & 34.0 & 79.1 & 87.6 & 41.2 & 72.6 & 76.6 & 65.0 & 26.9 & 69.4 & 82.8 & 40.7 & 55.9 & 81.1 & 72.9 & 62.7 & 81.4 & 53.6 & 43.2 & 65.6 &62.80\\ \hline
        \multicolumn{22}{l}{\textbf{hbox-level:}} \\ \hline
        H2RBox~\cite{yang2022h2rbox} &68.1 & 13.0 & 75.0 & 85.4 & 19.4 & 72.1 & 64.4 & 60.0 & 23.6 & 68.9 & 78.4 & 34.7 & 44.2 & 79.3 & 65.2 & 69.1 & 81.5 & 53.0 & 40.0 & 61.5 & 57.80\\
        H2RBox-v2~\cite{yu2023h2rbox} &67.2 & 37.7 & 55.6 & 80.8 & 29.3 & 66.8 & 76.1 & 58.4 & 26.4 & 53.9 & 80.3 & 25.3 & 48.9 & 78.8 & 67.6 & 62.4 & 82.5 & 49.7 & 42.0 & 63.1 & 57.64 \\ \hline
        \multicolumn{22}{l}{\textbf{PointOBB-based training:}} \\ \hline
        FCOS~\cite{tian2019fcos} & 58.4 & 17.1 & 70.7 & 77.7 & 0.1 & 70.3 & 64.7 & 4.5 & 7.2 & 0.8 & 74.2 & 9.9 & 9.1 & 69.0 & 38.2 & 49.8 & 46.1 & 16.8 & 32.4 & 29.6  & 37.31 \\
        Oriented R-CNN& 58.2 & 15.3 & 70.5 & 78.6 & 0.1 & 72.2 & 69.6 & 1.8 & 3.7 & 0.3 & 77.3 & 16.7 & 4.0 & 79.2 & 39.6 & 51.7 & 44.9 & 16.8 & 33.6 & 27.7 &38.08\\ \hline
        \multicolumn{22}{l}{\textbf{SAM-based training:}} \\ \hline
        RetinaNet~\cite{lin2017focal}&49.2&9.5&43.1&10.2&2.7&59.8&31.4&44.4&9.5&42.2&50.9&9.0&13.0&69.7&3.0&56.3&70.5&5.0&29.5&0.4&30.47\\
        FCOS~\cite{tian2019fcos} & 56.4&16.4&41.3&14.3&10.0&60.2&41.0&51.9&10.6&38.5&56.3&17.5&15.7&77.9&3.1&58.8&71.8&14.2&32.0&2.0&34.49 \\
        Oriented R-CNN~\cite{xie2021oriented}& 53.9&16.6&46.1&11.3&11.1&60.3&49.8&55.9&9.4&35.1&59.9&15.5&18.8&80.9&6.2&61.9&71.9&14.8&35.3&0.3 &35.75\\ \hline
        \multicolumn{22}{l}{\textbf{P2RBox-based training:}} \\ \hline
        \rowcolor{gray!20}RetinaNet& 52.9&10.6&67.1&47.6&5.3&66.2&54.3&43.5&10.0&61.0&64.8&20.0&17.5&70.2&44.1&58.2&78.9&11.3&32.3&8.2&41.21\\
        \rowcolor{gray!20}FCOS  & 59.5 & 20.8& 68.7 & 45.0 & 13.9& 68.7 & 58.8 & 56.8&15.7 &54.6 & 69.3 & 22.2 & 20.5 & 79.1 & 49.1 & 60.8 & 80.4 & 17.0 & 35.5 & 10.1  & 45.32 \\
        \rowcolor{gray!20}Oriented R-CNN & 62.1 & 22.4 & 70.3 & 54.2 & 15.3 & 69.5 & 64.7 & 60.4 & 15.9 & 53.2 & 72.5 & 22.6 & 26.4 & 80.9 & 50.2 & 62.2 & 80.9 & 19.4 & 38.8 & 3.9 &47.29\\  \bottomrule
        \end{tabular}
    }
\end{table*}

\subsection{Main Result}
\begin{figure}[htbp]
\begin{center}
\includegraphics[width=1\textwidth]{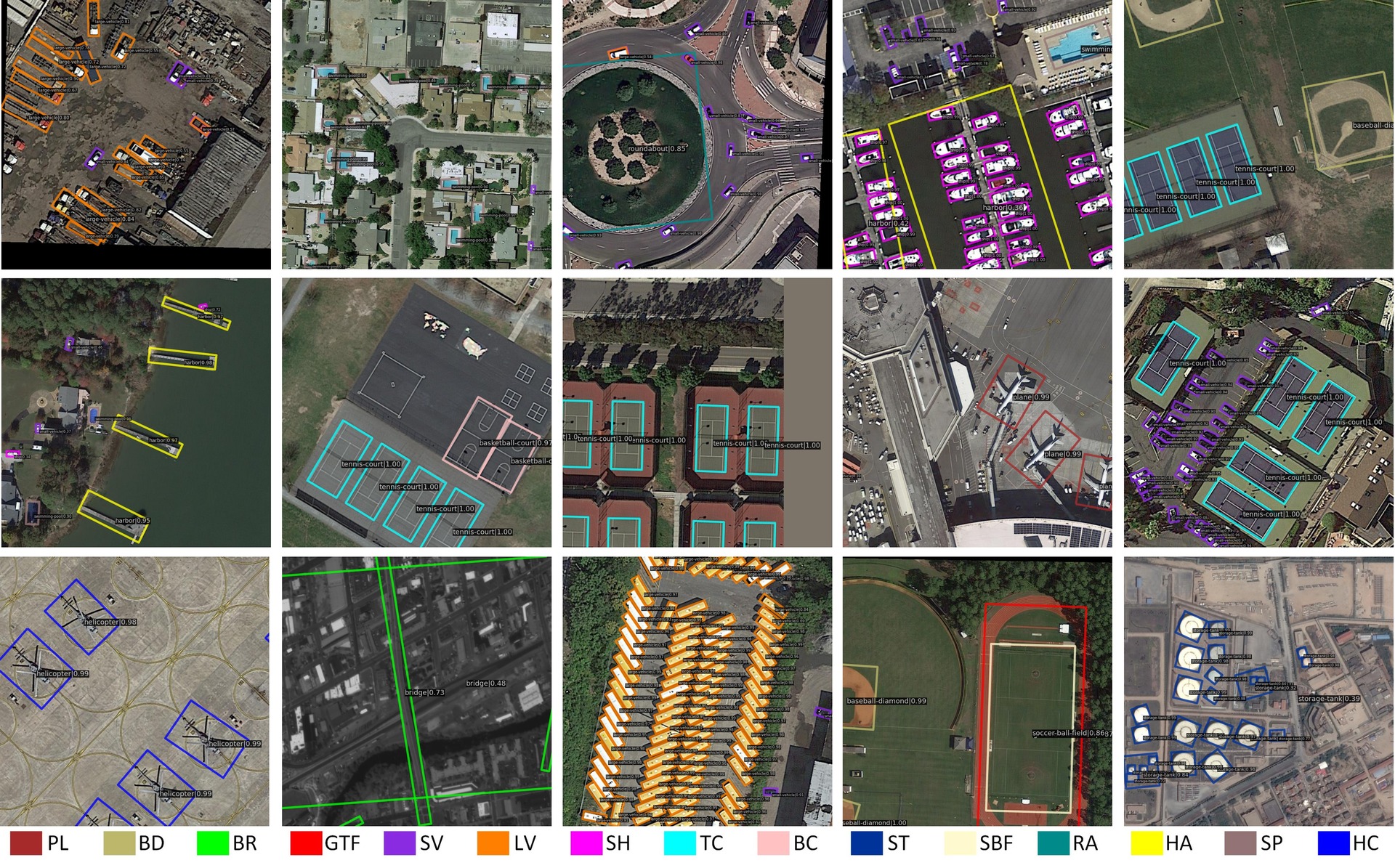}
\vspace{-12pt}
\caption{Examples of detection results on the DOTA dataset using P2RBox (Oriented R-CNN).}
\label{fig:visual}
\vspace{-10pt}
\end{center}
\end{figure}

As shown in Tab.\ref{table:dota}, our model's performance across many categories is obvious. In point-supervised detection, to demonstrate the effectiveness of the proposed method in our model, we designed a parameter-free rotation box annotation generator based on SAM, which directly retains the highest-score mask and computes the minimum bounding rectangle to obtain the rotated bounding box. By comparing the results of pseudo-label training on three strong supervised detectors, P2RBox model outperforms our baseline in every single category combined with any detector (56.66\% vs. 47.91\% on RetinaNet, 59.04\% vs. 50.84\% on FCOS, 62.43\% vs. 52.75\%  on Oriented R-CNN). 
The performance of the P2RBox network is significantly higher than previous point supervision methods (PointOBB 33.31\%, Point2RBox 40.27\%) , but there is still a considerable gap compared to strong supervision methods.
Examples of detection results on the DOTA dataset using P2RBox (Oriented R-CNN) are shown in Fig.~\ref{fig:visual}

The conclusions drawn from the DIOR-R dataset are similar to those from the DOTA dataset. P2RBox outperforms the parameter-free method based on SAM set in this paper (41.21\% vs. 30.47\% on RetinaNet, 45.32\% vs. 34.49\% on FCOS, 47.29\% vs. 35.75\% on Oriented R-CNN) . Under the same strong supervision detection training conditions, P2RBox verifies the high quality of the generated annotations compare to PointOBB (45.32\% vs. 37.31 on FCOS, 47.29\% vs. 38.08\% on Oriented R-CNN) .
\subsection{Ablation study}
According to Eq.~\ref{eq:score}, mask score can be calculated as follows:
\begin{equation}
    Score=S_{SAM}+S_{bmm}+S_{boxCls}-S_{offset}.
\end{equation}

\textbf{The impact of \modelname\ Modules.} 
As depicted in Tab.~\ref{tab:ab1}, each component in the model architecture was functionally verified and performance evaluated one by one. 
By using predicted angle or adding evaluation scores from different modules, this paper generates different rotation pseudo-box annotations and trains them on three mainstream detectors. The experimental results show good compatibility between the modules, and the use of each module contributes to the improvement of the final mask quality. 

\textbf{The impact of Mask quality assessment weights}.
For a candidate mask, $Score=S_{SAM}+\alpha \cdot S_{bmm}+\beta \cdot S_{cgm}$.
In this paper, the default weights between different metrics ($\alpha, \beta$)are set to 1. To test the impact of different weights on the results in the evaluation metrics of masks, Tab.~\ref{tab:ab2} designs experiments with varying weights for the first and second modules to demonstrate the stability of weights within a certain range. 
The final pseudo-rotation annotations generated by P2RBox and the pseudo-annotations generated by the SAM parameter-free method are compared with the ground truth (GT) in terms of mIoU in Tab.~\ref{tab:ab2.2}.


\begin{table}[htbp]
    \caption{The impact of each module of P2RBox on mAP$_50$}
    \begin{center}
    \resizebox{0.95\textwidth}{!}{
    \begin{tabular}{ccccc|ccc}
    \toprule
        \boldmath{$S_{SAM}$} & \boldmath{$S_{bmm}$} & \boldmath{$S_{boxCls}$} &\boldmath{$S_{offset}$}&\boldmath{$Angle$} & \textbf{RetinaNet} & \textbf{FCOS} & \textbf{Oriented R-CNN}  \\ \hline
    \checkmark & -& -&- &- &                                        47.91 & 50.84&52.75\\
    \checkmark & -&- &- & \checkmark&                              49.07 & 51.02&53.79\\
    \checkmark & \checkmark&- &- &\checkmark&                     49.60 & 52.68&55.45\\
    \checkmark & \checkmark&\checkmark &- &\checkmark&           52.35 & 54.73&58.00\\
    \checkmark & \checkmark&\checkmark &\checkmark &\checkmark& 56.66 & 59.04&62.43\\
    -& \checkmark&\checkmark &\checkmark &\checkmark&           56.78 & 59.35&62.00\\
    \bottomrule
    \end{tabular}}
    \label{tab:ab1}
    \end{center}
\end{table}

\begin{table}[htbp]
\begin{minipage}[t]{0.6\linewidth}
\caption{The impact of different $\alpha$ and $\beta$ on mIoU (DOTA)}
    \begin{center}
    \begin{tabular}{c|ccc}
    \toprule
    \diagbox{$\beta$}{$\alpha$} & 0.8 & 1.0 & 1.2    \\ \hline
    0.8 &60.64 &60.44 &60.13 \\
    1.0&60.65 &60.68& 60.49\\
    1.2&60.62 &60.60 &60.69\\ \bottomrule
    \end{tabular}
    \label{tab:ab2}
    \end{center}
    \vspace{-10pt}
\end{minipage}
\quad
\begin{minipage}[t]{0.36\linewidth}
    \caption{mIoU results on DOTA and DIOR-R using different methods.}
    \begin{center}
    \resizebox{0.9\textwidth}{!}{
    \begin{tabular}{c|cc}
    \toprule
    Method  & DOTA & DIOR-R   \\ \hline
    SAM & 54.57 & 49.23\\
    P2RBox & 60.68 & 59.61\\ \bottomrule
    \end{tabular}}
    \label{tab:ab2.2}
    \end{center}
    \vspace{-10pt}
\end{minipage}
\end{table}

\section{Conclusion}
This paper introduces P2RBox, the first SAM-based oriented object detector to our best knowledge. P2RBox distinguishes features through multi-instance learning, introduces a novel method for assessing proposal masks, designs a centrality-guided module for oriented bounding box conversion, and trains a fully supervised detector. P2RBox achieves impressive detection accuracy, with the exception of complex categories like BR. P2RBox offers a training paradigm that can be based on any proposal generator, and its generated rotated bounding box annotations can be used to train various strong supervised detectors, making it highly versatile and performance-adaptive without 
the need for additional parameters.

\clearpage

\newpage
\section*{Appendix}
\textbf{Upper Limit in our method}
In fact, we don't create new mask proposals; we simply choose a mask from the SAM generator using our criteria. As a result, there's a performance limit. When selecting based on IoU with the ground truth, the IoU results are displayed in Tab. \ref{tab:ap1}.
\begin{table}[htbp]
    \begin{minipage}[t]{1\linewidth}
    \caption{IoU result of SAM (highest score), P2RBox, ceiling (always choose the highest IoU using SAE on PL and HC while others minimum).}
    \vspace{-5pt}
    \begin{center}
    \resizebox{1\textwidth}{!}{
    \begin{tabular}{c|cccccccccccccccc}
    \hline
    Method  & PL & BD & BR & GTF& SV& LV& SH& TC& BC& ST& SBF & RA  & HA& SP& HC & mIoU  \\ \hline
    SAM&55.70&60.72&17.85&62.65&63.79&65.90&67.06&78.38&25.54&57.87&46.12&48.47&52.26&60.20&56.04&54.57\\
    P2RBox&71.22&66.10&22.01&64.83&65.42&69.22&67.97&80.70&44.80&58.49&66.95&52.22&57.30&63.50&59.54&60.68\\
    IoU-highest&74.08&70.39&26.23&78.53&69.61&73.48&74.91&83.43&47.14&64.61&70.08&58.37&66.51&66.81&64.11&65.89\\
    \end{tabular}}
    \label{tab:ap1}
    \end{center}
    \end{minipage}
    \vspace{-5pt}
\end{table}
This result demonstrates that we have outperformed the SAM model in every category compared to simply selecting the highest score. It also highlights that for some categories, the performance remains poor due to very low upper limits, despite significant improvements from the baseline.

\begin{figure}[htbp]
\begin{center}
\includegraphics[width=0.95\textwidth]{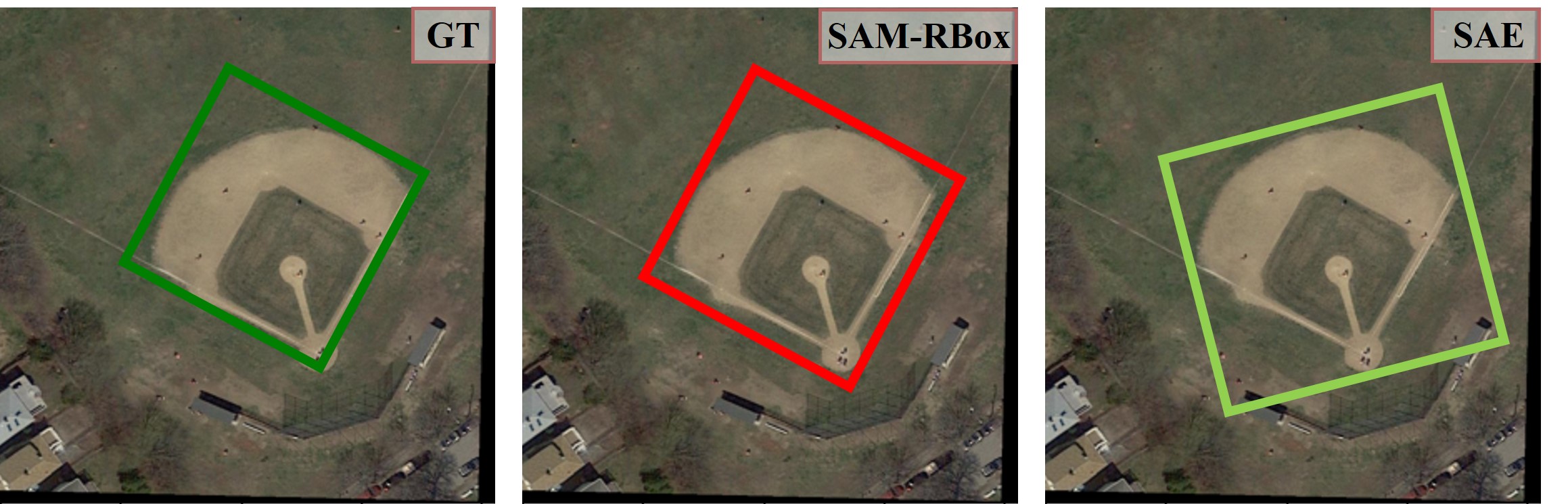}
\caption{GT, minimum and SAE on category BD.}
\label{fig:BD}
\end{center}
\end{figure}

\textbf{Details about the supervised value of angle}
Tab. \ref{tab:ap2} provides detailed information. The symmetric angle prediction method shows a slight decrease in IoU for some categories except for PL, which is negligible. However, it experiences a significant drop in the BD category.
\begin{table}[htbp]
    \begin{minipage}[t]{1\linewidth}
    \caption{Different mask2rbox method IoU results.}
    \begin{center}
    \resizebox{1\textwidth}{!}{
    \begin{tabular}{c|cccccccccccccccc}
    \hline
    Method  & PL & BD & BR & GTF& SV& LV& SH& TC& BC& ST& SBF & RA  & HA& SP& HC & mIoU  \\ \hline
    minimum-only&57.85&66.10&22.01&64.83&65.42&69.22&67.97&80.70&44.80&58.49&66.95&52.22&57.30&63.50&57.77&59.68\\
    supervised-value&71.22&58.14&21.80&64.94&65.46&69.12&68.15&80.37&43.80&56.00&64.91&52.87&57.42&62.95&59.54&59.78\\
    \end{tabular}}
    \label{tab:ap2}
    \end{center}
    \end{minipage}
    \vspace{-5pt}
\end{table}
The issue arises because the annotation or ground truth for BD does not align with its symmetry axis, even when a symmetry axis is present, as illustrated in Fig.~ \ref{fig:BD}.

\begin{figure}[htbp]
\begin{center}
\includegraphics[width=0.95\textwidth]{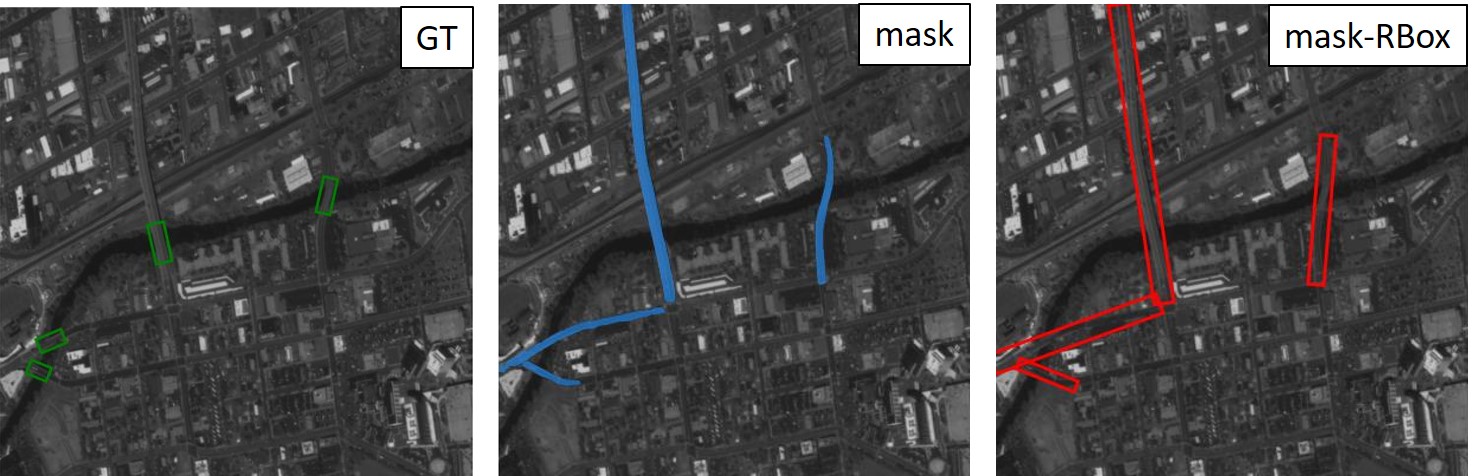}
\caption{Category BR, with mask proposals generated by SAM with annotated point and its circumscribed rbox.}
\label{fig:BR}
\end{center}
\end{figure}

\textbf{The limitations on the upper performance bound for the Bridges.}
category are quite restrictive. This is primarily attributed to the distinctive nature of its definition, which deviates from the conventional object definitions. In the case of bridges, they are defined as road segments that span across bodies of water, leading to a situation where there are insufficient discernible pixel variations between the left and right ends of the bridge. Consequently, this characteristic significantly hampers the performance of the SAM model. As a result, it imposes a notable constraint on the potential performance within this category. This challenge is further exemplified in Fig.~\ref{fig:BR}.

\begin{figure}[htbp]
\begin{center}
\includegraphics[width=0.95\textwidth]{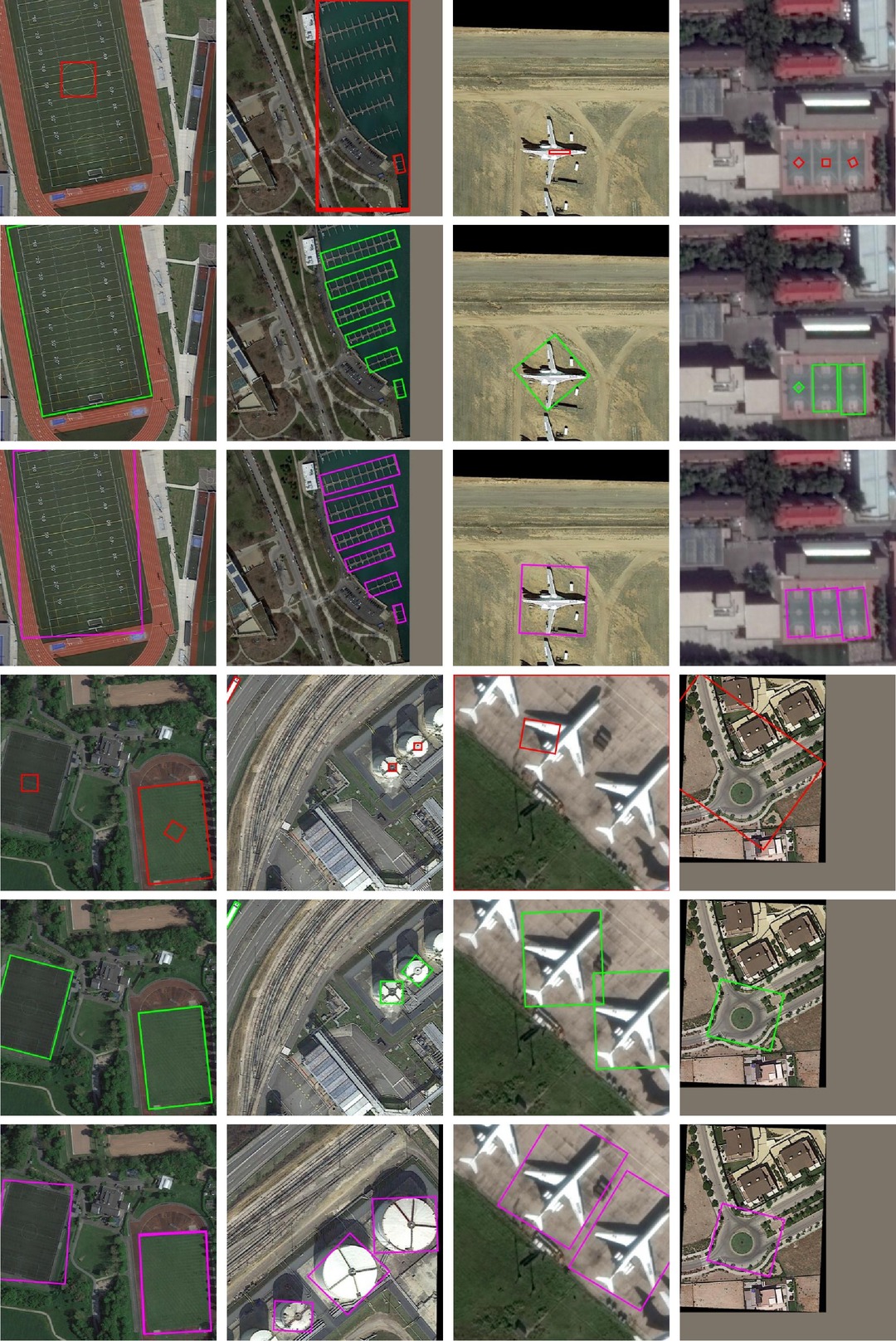}
\caption{Visualization of generated rbox with different approaches.}
\label{fig:vis1}
\end{center}
\end{figure}

\textbf{Further Visualizations of P2RBox}: Since P2RBox operates as a generative method, the characteristics of the rboxes it produces can provide insights into the model's behavior. Illustrated in Fig.~\ref{fig:vis1}, the red box represents the pseudo boxes directly generated by SAM, prioritizing the highest initial score while minimizing the bounding box. Conversely, the green section depicts $S_{SAM}+S_{sem}$ as the evaluation criteria. Additionally, the purple boxes are created with $Score=S_{SAM}+S_{bmm}+S_{cgm}$ and then transformed into rboxes with predicted angles aligned with the symmetric axis.

\begin{figure}[htbp]
\begin{center}
\includegraphics[width=0.95\textwidth]{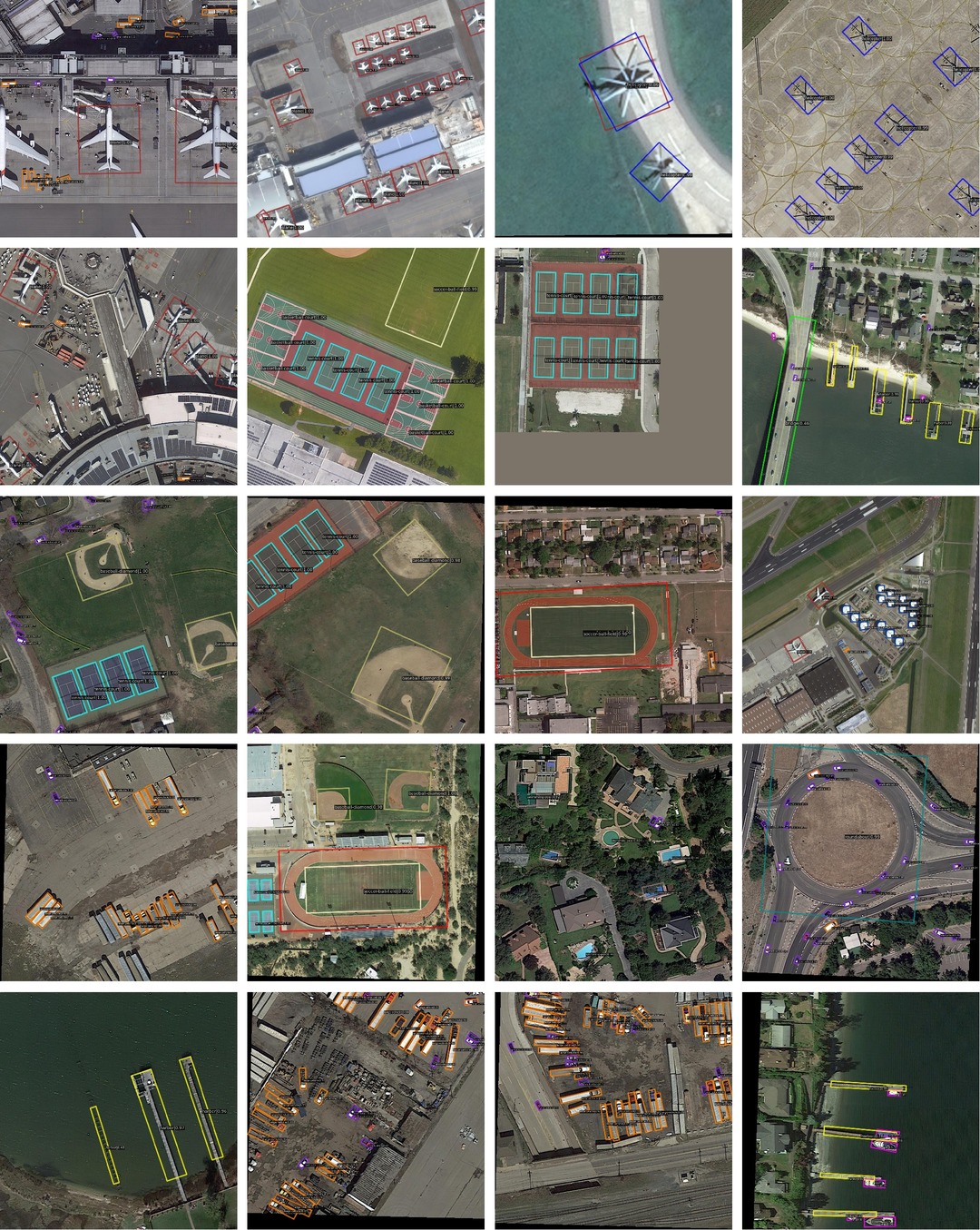}
\caption{Visualization of P2RBox (Oriented R-CNN).}
\label{fig:vis2}
\end{center}
\end{figure}

Fig.~\ref{fig:vis2} illustrates further insights into the inference outcomes on DOTA achieved with Oriented R-CNN trained on pseudo rboxes generated by P2RBox.

\begin{figure*}[htbp]
\begin{center}
\includegraphics[width=0.7\textwidth]{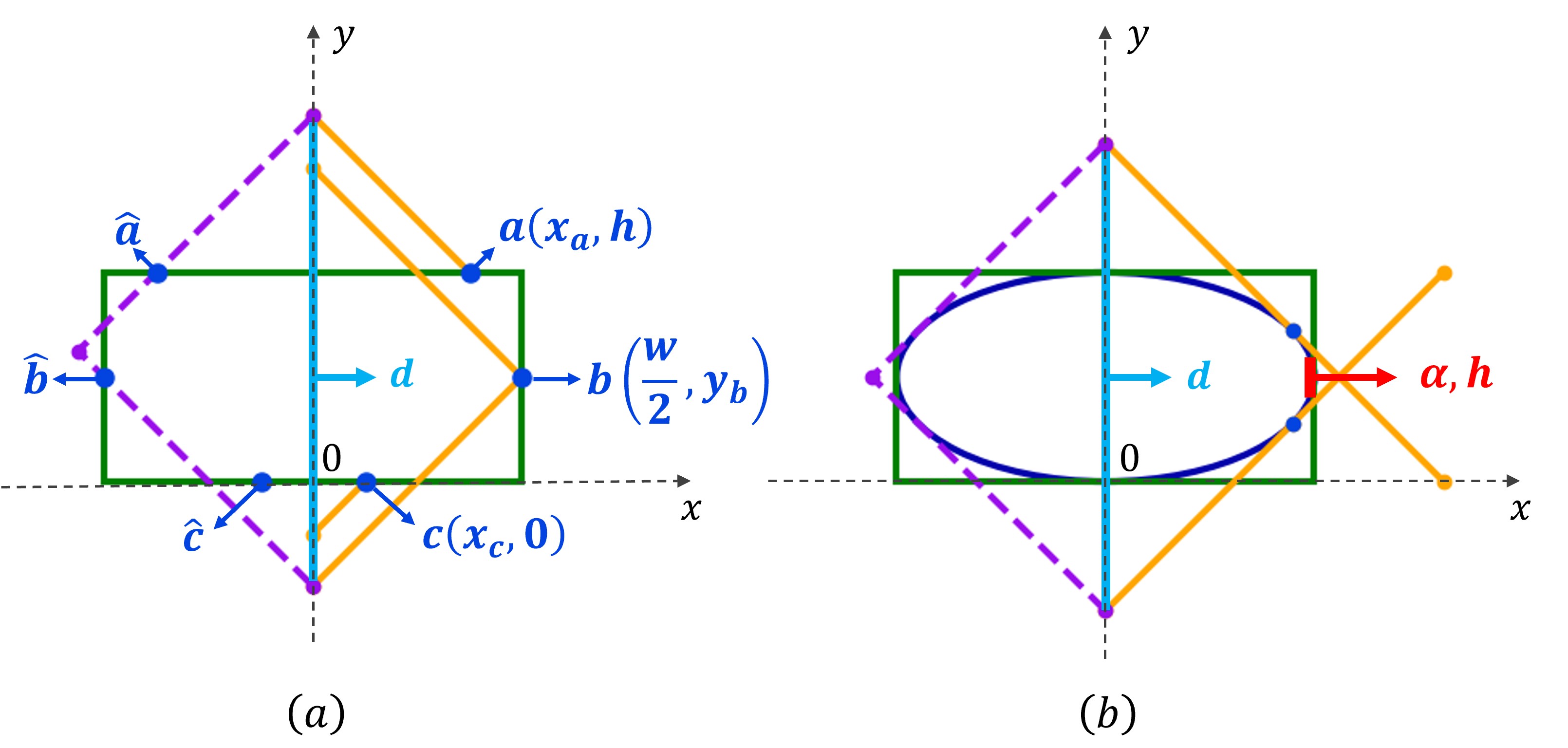}
\caption{Convex polygon example compared to the general case.}
\label{fig:pca}
\end{center}
\vspace{-10pt}
\end{figure*}

\textbf{What Result in Bad Bases using minimum bounding rectangle.} 
To illustrate this without loss of generality, let's consider an object that exhibits symmetry about the y-axis (see Fig. \ref{fig:pca}). We'll denote three points on the oriented circumscribed bounding box as $a$, $b$, and $c$, respectively, and their corresponding mirror points as $\hat{a}$, $\hat{b}$, and $\hat{c}$.

Now, suppose there exists a minimum circumscribed bounding box, denoted as $Mbox$, which is distinct from $Rbox$. By virtue of symmetry, $Mbox$ must also exhibit symmetry, compelling its shape to be square with its diagonal aligned along the axis of symmetry. To encompass the entire object, $Mbox$ must enclose points $a$, $b$, and $c$ as illustrated in Fig. \ref{fig:pca} (a). Let $d$ represent the length of the diagonal of $Mbox$. We have the following conditions:
\begin{equation}
    d >= \max(h + x_a, w/2 + y_b) - \min(-x_c, y_b - w/2),\\
\end{equation}
\begin{equation}
    d^2 <= 2 \times w \cdot h.
\end{equation}
The second inequality is derived based on the area requirement.
For the more general case, as shown in Fig. \ref{fig:pca} (b). By finding two tangent lines with fixed slopes (1 and -1), where $\alpha \cdot h$ is the distance between the intersections of these lines with the right green edge, we obtain an equation regarding the length of the diagonal:
\begin{equation}
    d = w + \alpha \cdot h.
\end{equation}
Specifically, if the width is equal to the height, $w=h$, the inequality simplifies to:
\begin{equation}
    \alpha <= \sqrt{2} - 1.
\end{equation}
In conclusion, taking an airplane as an example, as shown in the last column of Fig. \ref{fig:motivation}, due to the intersection ratio $\alpha < \sqrt{2}-1$, ambiguity arises between the minimum bounding rectangle and the oriented bounding rectangle, which is well addressed by Symmetry Axis Estimation Module.

\newpage

\clearpage
\bibliographystyle{unsrtnat}
\bibliography{neurips_2023}

\begin{thebibliography}{43}
\providecommand{\natexlab}[1]{#1}
\providecommand{\url}[1]{\texttt{#1}}
\expandafter\ifx\csname urlstyle\endcsname\relax
  \providecommand{\doi}[1]{doi: #1}\else
  \providecommand{\doi}{doi: \begingroup \urlstyle{rm}\Url}\fi

\bibitem[Ren et~al.(2015)Ren, He, Girshick, and Sun]{ren2015faster}
Shaoqing Ren, Kaiming He, Ross Girshick, and Jian Sun.
\newblock Faster r-cnn: Towards real-time object detection with region proposal networks.
\newblock In \emph{Advances in Neural Information Processing Systems}, pages 91--99, 2015.

\bibitem[Lin et~al.(2017)Lin, Goyal, Girshick, He, and Doll{\'a}r]{lin2017focal}
Tsung-Yi Lin, Priya Goyal, Ross Girshick, Kaiming He, and Piotr Doll{\'a}r.
\newblock Focal loss for dense object detection.
\newblock In \emph{Proceedings of the IEEE International Conference on Computer Vision}, pages 2980--2988, 2017.

\bibitem[Tian et~al.(2019)Tian, Shen, Chen, and He]{tian2019fcos}
Zhi Tian, Chunhua Shen, Hao Chen, and Tong He.
\newblock Fcos: Fully convolutional one-stage object detection.
\newblock In \emph{Proceedings of the IEEE International Conference on Computer Vision}, pages 9627--9636, 2019.

\bibitem[Ding et~al.(2019)Ding, Xue, Long, Xia, and Lu]{ding2019learning}
Jian Ding, Nan Xue, Yang Long, Gui-Song Xia, and Qikai Lu.
\newblock Learning roi transformer for oriented object detection in aerial images.
\newblock In \emph{Proceedings of the IEEE/CVF Conference on Computer Vision and Pattern Recognition}, pages 2849--2858, 2019.

\bibitem[Xie et~al.(2021)Xie, Cheng, Wang, Yao, and Han]{xie2021oriented}
Xingxing Xie, Gong Cheng, Jiabao Wang, Xiwen Yao, and Junwei Han.
\newblock Oriented r-cnn for object detection.
\newblock In \emph{Proceedings of the IEEE/CVF international conference on computer vision}, pages 3520--3529, 2021.

\bibitem[Bilen and Vedaldi(2016)]{bilen2016weakly}
Hakan Bilen and Andrea Vedaldi.
\newblock Weakly supervised deep detection networks.
\newblock In \emph{Proceedings of the IEEE conference on computer vision and pattern recognition}, pages 2846--2854, 2016.

\bibitem[Tang et~al.(2017)Tang, Wang, Bai, and Liu]{tang2017multiple}
Peng Tang, Xinggang Wang, Xiang Bai, and Wenyu Liu.
\newblock Multiple instance detection network with online instance classifier refinement.
\newblock In \emph{Proceedings of the IEEE conference on computer vision and pattern recognition}, pages 2843--2851, 2017.

\bibitem[Tang et~al.(2018)Tang, Wang, Bai, Shen, Bai, Liu, and Yuille]{tang2018pcl}
Peng Tang, Xinggang Wang, Song Bai, Wei Shen, Xiang Bai, Wenyu Liu, and Alan Yuille.
\newblock Pcl: Proposal cluster learning for weakly supervised object detection.
\newblock \emph{IEEE transactions on pattern analysis and machine intelligence}, 42\penalty0 (1):\penalty0 176--191, 2018.

\bibitem[Chen et~al.(2020)Chen, Fu, Jiang, Chen, and Hua]{chen2020slv}
Ze~Chen, Zhihang Fu, Rongxin Jiang, Yaowu Chen, and Xian-Sheng Hua.
\newblock Slv: Spatial likelihood voting for weakly supervised object detection.
\newblock In \emph{Proceedings of the IEEE/CVF Conference on Computer Vision and Pattern Recognition}, pages 12995--13004, 2020.

\bibitem[Wan et~al.(2018)Wan, Wei, Jiao, Han, and Ye]{wan2018min}
Fang Wan, Pengxu Wei, Jianbin Jiao, Zhenjun Han, and Qixiang Ye.
\newblock Min-entropy latent model for weakly supervised object detection.
\newblock In \emph{Proceedings of the IEEE conference on computer vision and pattern recognition}, pages 1297--1306, 2018.

\bibitem[Zhou et~al.(2016)Zhou, Khosla, Lapedriza, Oliva, and Torralba]{zhou2016learning}
Bolei Zhou, Aditya Khosla, Agata Lapedriza, Aude Oliva, and Antonio Torralba.
\newblock Learning deep features for discriminative localization.
\newblock In \emph{Proceedings of the IEEE conference on computer vision and pattern recognition}, pages 2921--2929, 2016.

\bibitem[Diba et~al.(2017)Diba, Sharma, Pazandeh, Pirsiavash, and Van~Gool]{diba2017weakly}
Ali Diba, Vivek Sharma, Ali Pazandeh, Hamed Pirsiavash, and Luc Van~Gool.
\newblock Weakly supervised cascaded convolutional networks.
\newblock In \emph{Proceedings of the IEEE conference on computer vision and pattern recognition}, pages 914--922, 2017.

\bibitem[Zhang et~al.(2018)Zhang, Wei, Feng, Yang, and Huang]{zhang2018adversarial}
Xiaolin Zhang, Yunchao Wei, Jiashi Feng, Yi~Yang, and Thomas~S Huang.
\newblock Adversarial complementary learning for weakly supervised object localization.
\newblock In \emph{Proceedings of the IEEE conference on computer vision and pattern recognition}, pages 1325--1334, 2018.

\bibitem[Tan et~al.(2023)Tan, Jiang, Guo, and Zhang]{tan2023wsodet}
Zhiwen Tan, Zhiguo Jiang, Chen Guo, and Haopeng Zhang.
\newblock Wsodet: A weakly supervised oriented detector for aerial object detection.
\newblock \emph{IEEE Transactions on Geoscience and Remote Sensing}, 61:\penalty0 1--12, 2023.

\bibitem[Yang et~al.(2022)Yang, Zhang, Li, Wang, Zhou, and Yan]{yang2022h2rbox}
Xue Yang, Gefan Zhang, Wentong Li, Xuehui Wang, Yue Zhou, and Junchi Yan.
\newblock H2rbox: Horizontal box annotation is all you need for oriented object detection.
\newblock \emph{arXiv preprint arXiv:2210.06742}, 2022.

\bibitem[Yu et~al.(2024)Yu, Yang, Li, Zhou, Da, and Yan]{yu2024h2rbox}
Yi~Yu, Xue Yang, Qingyun Li, Yue Zhou, Feipeng Da, and Junchi Yan.
\newblock H2rbox-v2: Incorporating symmetry for boosting horizontal box supervised oriented object detection.
\newblock \emph{Advances in Neural Information Processing Systems}, 36, 2024.

\bibitem[Yi et~al.(2023)Yi, Yang, Li, Da, Yan, Dai, and Qiao]{yi2023point2rbox}
Yu~Yi, Xue Yang, Qingyun Li, Feipeng Da, Junchi Yan, Jifeng Dai, and Yu~Qiao.
\newblock Point2rbox: Combine knowledge from synthetic visual patterns for end-to-end oriented object detection with single point supervision.
\newblock \emph{arXiv preprint arXiv:2311.14758}, 2023.

\bibitem[Luo et~al.(2023)Luo, Yang, Yu, Li, Yan, and Li]{luo2023pointobb}
Junwei Luo, Xue Yang, Yi~Yu, Qingyun Li, Junchi Yan, and Yansheng Li.
\newblock Pointobb: Learning oriented object detection via single point supervision.
\newblock \emph{arXiv preprint arXiv:2311.14757}, 2023.

\bibitem[Kirillov et~al.(2023)Kirillov, Mintun, Ravi, Mao, Rolland, Gustafson, Xiao, Whitehead, Berg, Lo, et~al.]{kirillov2023segment}
Alexander Kirillov, Eric Mintun, Nikhila Ravi, Hanzi Mao, Chloe Rolland, Laura Gustafson, Tete Xiao, Spencer Whitehead, Alexander~C Berg, Wan-Yen Lo, et~al.
\newblock Segment anything.
\newblock \emph{arXiv preprint arXiv:2304.02643}, 2023.

\bibitem[Yang et~al.(2021)Yang, Yan, Feng, and He]{yang2021r3det}
Xue Yang, Junchi Yan, Ziming Feng, and Tao He.
\newblock R3det: Refined single-stage detector with feature refinement for rotating object.
\newblock In \emph{Proceedings of the AAAI Conference on Artificial Intelligence}, volume~35, pages 3163--3171, 2021.

\bibitem[Han et~al.(2021{\natexlab{a}})Han, Ding, Li, and Xia]{han2021align}
Jiaming Han, Jian Ding, Jie Li, and Gui-Song Xia.
\newblock Align deep features for oriented object detection.
\newblock \emph{IEEE Transactions on Geoscience and Remote Sensing}, 60:\penalty0 1--11, 2021{\natexlab{a}}.

\bibitem[Han et~al.(2021{\natexlab{b}})Han, Ding, Xue, and Xia]{han2021redet}
Jiaming Han, Jian Ding, Nan Xue, and Gui-Song Xia.
\newblock Redet: A rotation-equivariant detector for aerial object detection.
\newblock In \emph{Proceedings of the IEEE Conference on Computer Vision and Pattern Recognition}, pages 2786--2795, 2021{\natexlab{b}}.

\bibitem[Iqbal et~al.(2021)Iqbal, Munir, Mahmood, Ali, and Ali]{iqbal2021leveraging}
Javed Iqbal, Muhammad~Akhtar Munir, Arif Mahmood, Afsheen~Rafaqat Ali, and Mohsen Ali.
\newblock Leveraging orientation for weakly supervised object detection with application to firearm localization.
\newblock \emph{Neurocomputing}, 440:\penalty0 310--320, 2021.

\bibitem[Yu et~al.(2023)Yu, Yang, Li, Zhou, Zhang, Yan, and Da]{yu2023h2rbox}
Yi~Yu, Xue Yang, Qingyun Li, Yue Zhou, Gefan Zhang, Junchi Yan, and Feipeng Da.
\newblock H2rbox-v2: Boosting hbox-supervised oriented object detection via symmetric learning.
\newblock \emph{arXiv preprint arXiv:2304.04403}, 2023.

\bibitem[Zhu et~al.(2023)Zhu, Ferenczi, Purkait, Drummond, Rezatofighi, and Van Den~Hengel]{zhu2023knowledge}
Tianyu Zhu, Bryce Ferenczi, Pulak Purkait, Tom Drummond, Hamid Rezatofighi, and Anton Van Den~Hengel.
\newblock Knowledge combination to learn rotated detection without rotated annotation.
\newblock In \emph{Proceedings of the IEEE/CVF Conference on Computer Vision and Pattern Recognition}, pages 15518--15527, 2023.

\bibitem[Hua et~al.(2023)Hua, Liang, Li, Liu, Zou, Ye, and Bai]{hua2023sood}
Wei Hua, Dingkang Liang, Jingyu Li, Xiaolong Liu, Zhikang Zou, Xiaoqing Ye, and Xiang Bai.
\newblock Sood: Towards semi-supervised oriented object detection.
\newblock In \emph{Proceedings of the IEEE/CVF Conference on Computer Vision and Pattern Recognition}, pages 15558--15567, 2023.

\bibitem[Zhou et~al.(2023)Zhou, Jiang, Chen, Chen, and Liu]{zhou2023semi}
Yue Zhou, Xue Jiang, Zeming Chen, Lin Chen, and Xingzhao Liu.
\newblock A semi-supervised arbitrary-oriented sar ship detection network based on interference consistency learning and pseudo label calibration.
\newblock \emph{IEEE Journal of Selected Topics in Applied Earth Observations and Remote Sensing}, 2023.

\bibitem[Wu et~al.(2024)Wu, Wong, and Wu]{wu2024pseudo}
Wenhao Wu, Hau-San Wong, and Si~Wu.
\newblock Pseudo-siamese teacher for semi-supervised oriented object detection.
\newblock \emph{IEEE Transactions on Geoscience and Remote Sensing}, 2024.

\bibitem[Chan et~al.(2024)Chan, Sahni, Li, Luthra, Fang, Pouch, and Rajapakse]{chan2024sam3d}
Trevor~J Chan, Aarush Sahni, Jie Li, Alisha Luthra, Amy Fang, Alison Pouch, and Chamith~S Rajapakse.
\newblock Sam3d: Zero-shot semi-automatic segmentation in 3d medical images with the segment anything model.
\newblock \emph{arXiv preprint arXiv:2405.06786}, 2024.

\bibitem[Zohranyan et~al.(2024)Zohranyan, Navasardyan, Navasardyan, Borggrefe, and Navasardyan]{zohranyan2024dr}
Vazgen Zohranyan, Vagner Navasardyan, Hayk Navasardyan, Jan Borggrefe, and Shant Navasardyan.
\newblock Dr-sam: An end-to-end framework for vascular segmentation, diameter estimation, and anomaly detection on angiography images.
\newblock \emph{arXiv preprint arXiv:2404.17029}, 2024.

\bibitem[Li et~al.(2024{\natexlab{a}})Li, Qi, Yu, Huo, Shi, and Gao]{li2024concatenate}
Shumeng Li, Lei Qi, Qian Yu, Jing Huo, Yinghuan Shi, and Yang Gao.
\newblock Concatenate, fine-tuning, re-training: A sam-enabled framework for semi-supervised 3d medical image segmentation.
\newblock \emph{arXiv preprint arXiv:2403.11229}, 2024{\natexlab{a}}.

\bibitem[Liu et~al.(2024)Liu, Yang, van Diest, Pluim, and Veta]{liu2024wsi}
Hong Liu, Haosen Yang, Paul~J van Diest, Josien~PW Pluim, and Mitko Veta.
\newblock Wsi-sam: Multi-resolution segment anything model (sam) for histopathology whole-slide images.
\newblock \emph{arXiv preprint arXiv:2403.09257}, 2024.

\bibitem[Chen et~al.(2024)Chen, Xu, Liu, and Yuan]{chen2024sam}
Zhen Chen, Qing Xu, Xinyu Liu, and Yixuan Yuan.
\newblock Un-sam: Universal prompt-free segmentation for generalized nuclei images.
\newblock \emph{arXiv preprint arXiv:2402.16663}, 2024.

\bibitem[Wei et~al.(2023)Wei, Chen, Yu, Li, Jiao, and Han]{wei2023semantic}
Zhaoyang Wei, Pengfei Chen, Xuehui Yu, Guorong Li, Jianbin Jiao, and Zhenjun Han.
\newblock Semantic-aware sam for point-prompted instance segmentation.
\newblock \emph{arXiv preprint arXiv:2312.15895}, 2023.

\bibitem[Zhang et~al.(2024{\natexlab{a}})Zhang, Liu, Lin, Liao, and Li]{zhang2024uv}
Xin Zhang, Yu~Liu, Yuming Lin, Qingmin Liao, and Yong Li.
\newblock Uv-sam: Adapting segment anything model for urban village identification.
\newblock In \emph{Proceedings of the AAAI Conference on Artificial Intelligence}, volume~38, pages 22520--22528, 2024{\natexlab{a}}.

\bibitem[Li et~al.(2024{\natexlab{b}})Li, Xiao, and Tang]{li2024asam}
Bo~Li, Haoke Xiao, and Lv~Tang.
\newblock Asam: Boosting segment anything model with adversarial tuning.
\newblock \emph{arXiv preprint arXiv:2405.00256}, 2024{\natexlab{b}}.

\bibitem[Zhang et~al.(2024{\natexlab{b}})Zhang, Yan, Liu, and Lu]{zhang2024fantastic}
Pingping Zhang, Tianyu Yan, Yang Liu, and Huchuan Lu.
\newblock Fantastic animals and where to find them: Segment any marine animal with dual sam.
\newblock \emph{arXiv preprint arXiv:2404.04996}, 2024{\natexlab{b}}.

\bibitem[Chen et~al.(2022)Chen, Yu, Han, Hassan, Wang, Li, Zhao, Shi, Han, and Ye]{chen2022point}
Pengfei Chen, Xuehui Yu, Xumeng Han, Najmul Hassan, Kai Wang, Jiachen Li, Jian Zhao, Humphrey Shi, Zhenjun Han, and Qixiang Ye.
\newblock Point-to-box network for accurate object detection via single point supervision.
\newblock In \emph{European Conference on Computer Vision}, pages 51--67. Springer, 2022.

\bibitem[Xia et~al.(2018)Xia, Bai, Ding, Zhu, Belongie, Luo, Datcu, Pelillo, and Zhang]{xia2018dota}
Gui-Song Xia, Xiang Bai, Jian Ding, Zhen Zhu, Serge Belongie, Jiebo Luo, Mihai Datcu, Marcello Pelillo, and Liangpei Zhang.
\newblock Dota: A large-scale dataset for object detection in aerial images.
\newblock In \emph{Proceedings of the IEEE Conference on Computer Vision and Pattern Recognition}, pages 3974--3983, 2018.

\bibitem[Zhou et~al.(2022)Zhou, Yang, Zhang, Wang, Liu, Hou, Jiang, Liu, Yan, Lyu, Zhang, and Chen]{zhou2022mmrotate}
Yue Zhou, Xue Yang, Gefan Zhang, Jiabao Wang, Yanyi Liu, Liping Hou, Xue Jiang, Xingzhao Liu, Junchi Yan, Chengqi Lyu, Wenwei Zhang, and Kai Chen.
\newblock Mmrotate: A rotated object detection benchmark using pytorch.
\newblock In \emph{Proceedings of the 30th ACM International Conference on Multimedia}, page 7331–7334, 2022.

\bibitem[Cheng et~al.(2022)Cheng, Wang, Li, Xie, Lang, Yao, and Han]{dior-r}
Gong Cheng, Jiabao Wang, Ke~Li, Xingxing Xie, Chunbo Lang, Yanqing Yao, and Junwei Han.
\newblock Anchor-free oriented proposal generator for object detection.
\newblock \emph{IEEE Transactions on Geoscience and Remote Sensing}, 60:\penalty0 1--11, 2022.

\bibitem[Li et~al.(2020)Li, Wan, Cheng, Meng, and Han]{dior}
Ke~Li, Gang Wan, Gong Cheng, Liqiu Meng, and Junwei Han.
\newblock Object detection in optical remote sensing images: A survey and a new benchmark.
\newblock \emph{ISPRS journal of photogrammetry and remote sensing}, 159:\penalty0 296--307, 2020.

\bibitem[Bottou(2012)]{bottou2012stochastic}
L{\'e}on Bottou.
\newblock Stochastic gradient descent tricks.
\newblock In \emph{Neural Networks: Tricks of the Trade: Second Edition}, pages 421--436. Springer, 2012.

\end{thebibliography}

\clearpage

\end{document}